\documentclass[journal]{IEEEtran}
\usepackage[T1]{fontenc}
\usepackage{newtxtext,newtxmath}
\usepackage{graphicx,booktabs,array,multirow,amsmath}
\usepackage{cite,placeins}
\usepackage[hidelinks]{hyperref}
\graphicspath{{figures/}}

\newcommand{\R}{\mathbb{R}}
\title{Task Adaptation Through Guidance in Frozen Feature Upsampling}
\author{Pingan~Wang%
\thanks{Pingan Wang is with the College of Artificial Intelligence and Low-Altitude Technology, South China Agricultural University, Guangzhou 510642, China (e-mail: pinganwang@scau.edu.cn). Corresponding author: Pingan Wang.}}
\hypersetup{pdfauthor={Pingan Wang},pdftitle={Task Adaptation Through Guidance in Frozen Feature Upsampling}}

\newif\ifQFourmeasureOverview
\makeatletter
\let\QFourEndFloatBox\@endfloatbox
\def\@endfloatbox{\QFourEndFloatBox
  \ifQFourmeasureOverview
    \dimen@=\ht\@currbox\advance\dimen@ by\dp\@currbox
    \advance\dimen@ by\textfloatsep
    \typeout{Q04_OVERVIEW_WITH_CAPTION_AND_SPACING_PT=\the\dimen@; HALF_TEXT_HEIGHT_PT=\the\dimexpr.5\textheight\relax}
    \ifdim\dimen@>.5\textheight
      \PackageError{Q04 figure layout}{Overview exceeds half a page}{Reflow or reduce the overview width.}
    \fi
    \global\QFourmeasureOverviewfalse
  \fi}
\makeatother

\begin{document}
\raggedbottom
\bstctlcite{IEEEcontrol}
\maketitle
\begin{abstract}
Adapting features before upsampling can improve dense prediction, but the gains may arise from changes in feature values, spatial aggregation, or both. We investigate this ambiguity by comparing matched residual adapters before and after a frozen JAFAR upsampler under a shared parent. We construct a readout-null intervention that preserves a frozen linear classifier's coarse logits while allowing spatial attention weights to adapt. Guidance/value exchanges on trained models and retrained fixed-guidance controls test the utility of this adaptation route. In three-seed experiments on PASCAL Visual Object Classes (VOC) and Cityscapes, adaptation before upsampling improves mean intersection over union (mIoU) by 0.807 and 1.413 percentage points, respectively, over adaptation after upsampling. Across both datasets and the JAFAR and AnyUp upsamplers, readout-null adaptation produces mean gains of 1.379--1.987 points over the corresponding frozen linear baselines. Retrained controls further support the benefit of adaptive spatial guidance. These findings explain how dense predictions can improve despite unchanged coarse classifier outputs and identify spatial guidance as a useful target when adapting frozen feature upsampling systems.
\end{abstract}

\begin{IEEEkeywords}
Feature upsampling, semantic segmentation, frozen models, parameter-efficient adaptation, controlled experiments.
\end{IEEEkeywords}
\section{Introduction}
\begin{figure*}[t]
\centering\includegraphics[width=\textwidth]{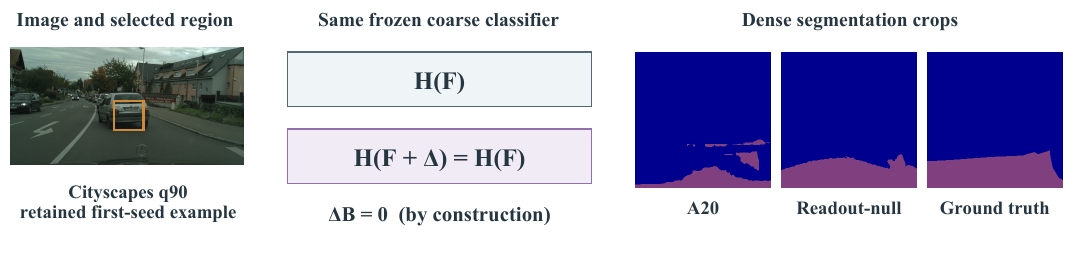}
\caption{A fixed coarse readout need not fix the dense prediction. The equality is a schematic consequence of $\Delta B=0$; the masks are retained JAFAR predictions for the original Cityscapes q90 example, revisited in Fig.~\ref{fig:cases}.}
\label{fig:problem}
\end{figure*}

\IEEEPARstart{D}{ense} visual prediction requires semantic representations at a resolution finer than the token grid produced by a vision transformer. With self-supervised representations, a compact task head can exploit frozen semantic features while an upsampler supplies spatial detail for pixelwise decisions. Vision transformers and self-supervised encoders have made frozen features an effective starting point for this design~\cite{dosovitskiy2021vit,caron2021dino,oquab2024dinov2}. Feature upsampling methods now provide learned, image-guided mappings from a coarse grid to a dense field~\cite{fu2024featup,couairon2025jafar,ranzinger2025featsharp,wimmer2026anyup}.

Where should a small task adapter enter this system? Fig.~\ref{fig:problem} makes the question concrete: a residual can preserve every coarse classifier logit while changing the dense segmentation. Applying it after upsampling changes the dense feature values presented to the task head. Applying the same adapter before upsampling can also change the rule that selects and combines coarse tokens. The distinction matters when the upsampler conditions its spatial weights on the input features. A pre-upsampling gain can therefore involve spatial guidance, feature values, or their interaction. We therefore ask whether adapting spatial guidance can improve dense prediction while the coarse classifier outputs remain fixed.

We investigate the hypothesis that feature-conditioned spatial guidance provides a useful route for task adaptation. We first examine it in a fixed DINOv2--JAFAR segmentation system, then test the readout-null and guidance-retraining results with a second frozen operator, AnyUp. We place identically initialized residual adapters before or after a shared frozen upsampler and nonlinear parent, making placement the controlled architectural difference. We then construct a more restrictive instrument, which we call \emph{readout-null adaptation}. For a frozen linear classifier with weight matrix $B$, we constrain the added residual $\Delta$ to satisfy $\Delta B=0$. The adapted feature therefore has exactly the same coarse classifier logits in exact arithmetic. JAFAR can still assign different spatial weights to the adapted feature. We use this construction to test whether spatial guidance can improve dense predictions while the coarse classifier output stays fixed.

The readout-null constraint has a useful algebraic consequence. If the upsampling weights are held fixed, adding $\Delta$ to the values cannot change the final linear readout. If the adapted weights are retained but the values are restored to their original features, the final readout is again unchanged. We test both implications by intervening on already trained checkpoints. The exchanges check the model-specific readout identities on the learned checkpoints. We separately retrain a fixed-guidance control to test whether allowing guidance to adapt is useful during optimization. The exchanges check the learned forward function; retraining measures the performance attainable under a fixed guidance rule.

Across three new training seeds, we find a pre-upsampling advantage on VOC and on a Cityscapes configuration. The mean paired differences are $0.807$ and $1.413$ percentage points (pp). Readout-null adaptation exceeds the frozen linear parent by $1.402$ and $1.987$ pp. Each dataset has its own frozen two-comparison family; the reported primary intervals account for these two comparisons. We address complementary explanations with a larger frozen nonlinear parent, fixed-guidance retraining, numerical guidance/value exchanges, and a retrospective equivalence analysis.

The experiments build one argument: locate the useful adaptation, hold the coarse readout fixed, and test whether allowing guidance to adapt remains useful after retraining and transfer. We examine it with matched residuals, full-validation checkpoint exchanges, and controlled comparisons on two datasets. In the independent 24-fit AnyUp extension, we recover both the readout-null gain and the utility of adaptive guidance on both datasets; all four comparisons pass their separately defined family. Together, the experiments characterize a shared adaptation route in the two frozen upsamplers.

\section{Related Work}
\subsection{Dense prediction from frozen representations}
Classical fully convolutional and encoder--decoder segmentation systems learn dense representations directly~\cite{long2015fcn,chen2018deeplabv3p}. Transformer systems provide alternative encoder and decoder designs~\cite{strudel2021segmenter,xie2021segformer,cheng2022mask2former}. In parallel, DINO, iBOT, masked autoencoding, and DINOv2 have strengthened general-purpose visual representations~\cite{caron2021dino,zhou2022ibot,he2022mae,oquab2024dinov2}. Our setting keeps the representation and upsampler frozen and studies how a small supervised component uses them. We isolate adapter placement within a fixed segmentation system.

\subsection{Guided feature upsampling}
The separation between a signal to be aggregated and information that guides aggregation predates neural feature upsampling. Bilateral filtering, joint bilateral upsampling, and guided image filtering provide relevant examples~\cite{tomasi1998bilateral,kopf2007jbu,he2010guided}. Recent feature upsamplers extend this design space with learned and attention-based operators. FeatUp provides a model-agnostic resolution framework; JAFAR, FeatSharp, LoftUp, and AnyUp develop complementary ways to recover spatially detailed features~\cite{fu2024featup,couairon2025jafar,ranzinger2025featsharp,huang2025loftup,wimmer2026anyup}. Neighborhood Attention Filtering, Upsample Anything, and PixelUp extend zero-shot feature upsampling with additional operator designs~\cite{chambon2025naf,seo2025upa,singh2026pixelup}. We hold JAFAR fixed within the placement study. In the independent AnyUp extension, we change the operator and repeat the null and guidance controls within it. This design compares adaptation routes under each frozen operator.

AnyUp already ablates the feature-to-key information path and finds a useful image-derived spatial prior without that path~\cite{wimmer2026anyup}. That experiment studies an upsampler component, whereas we freeze the upsampler and restrict a supervised task residual to the classifier null space. We make feature-conditioned spatial attention an explicit architectural condition of the argument. For example, RaysUp describes image-derived geometric guidance with target features serving as values~\cite{ding2026raysup}. Under this affine-readout factorization, a readout-null residual leaves dense logits unchanged when the spatial weights are independent of it. Our experiments examine the feature-conditioned case. 

\subsection{Task adaptation and null space instruments}
Low-rank adaptation, transformer adapters, convolutional bypasses, and visual adapters provide established routes for adapting large representations efficiently~\cite{hu2022lora,chen2022adaptformer,jie2022convpass,yin2025mona}. Task-specific clues and semantic or visual context also motivate lightweight dense-prediction adaptation~\cite{chen2026clues,wang2026sav}. EASe combines feature calibration with self-supervised upsampling~\cite{singh2026ease}. We study the adaptation route under a common frozen parent, matched residual initialization, and paired data streams.

Null space projection and linear concept erasure restrict what a linear readout can detect~\cite{ravfogel2020inlp,belrose2023leace}. We project only the added residual and retain the original feature as the skip connection. The resulting identities are consequences of the frozen affine readout and shared linear value aggregation. The empirical question is whether this constrained residual can still improve dense segmentation, and whether the observed implementation respects those identities. The intervention preserves the coarse logits of one fixed classifier by constraining the added residual.

\section{Hypothesis and Test Design}

\subsection{Locating adaptation in the frozen system}
Let $F\in\R^{N\times D}$ be coarse image features and $I$ the image. For the reference operator used here, dense features can be written as
\begin{equation}
 U_I(F)=A(I,F)F,\qquad A(I,F)\mathbf1_N=\mathbf1_M.
 \label{eq:upsampling}
\end{equation}
The matrix $A$ represents the spatial attention weights after averaging attention heads. JAFAR obtains queries from the image and conditions keys on the coarse features through a Spatial Feature Transform (SFT), following the terminology of its architecture~\cite{couairon2025jafar}. Its weights aggregate the unprojected feature values. We use this factorization for the pinned reference implementation.

For a frozen task head $H$ and a residual map $T(F)=F+R(F)$, ordinary pre- and post-upsampling adaptation have logits
\begin{align}
 L_{\rm Pre}&=H\!\left(U_I(T(F))\right),\\
 L_{\rm Post}&=H\!\left(T(U_I(F))\right).
\end{align}
For each feature vector $f$, the residual increment is
\begin{equation}
 r(f)=W_2\operatorname{GELU}\!\left(W_1\frac{f}{\|f\|_2+10^{-6}}+b_1\right)+b_2.
\end{equation}
We initialize $W_2$ and $b_2$ to zero.

\begin{figure}[t]
\centering\includegraphics[width=\columnwidth]{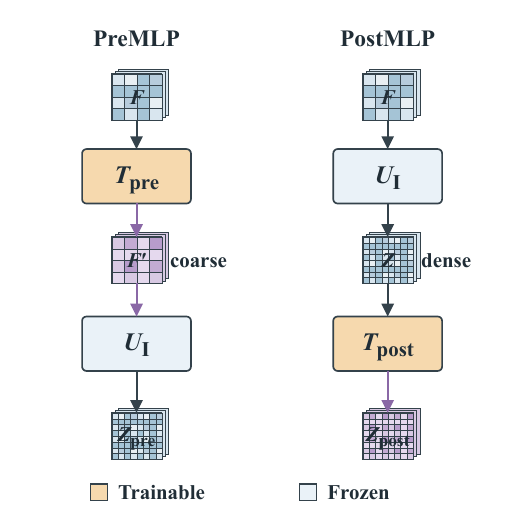}
\caption{Local placement of the pointwise residual relative to the frozen image-conditioned upsampler $U_I$. PreMLP adapts coarse features and PostMLP adapts dense features; both use a common frozen nonlinear parent. Grids are schematic.}
\label{fig:placement}
\end{figure}

We use the same pointwise residual architecture on either grid. In the placement experiment, we freeze a common nonlinear parent head and pair the initial residual parameters, images, and random flips. This pairing makes adapter placement the controlled architectural difference. Fig.~\ref{fig:placement} isolates this ordering; the two residuals are optimized separately from the paired initial state.

\subsection{Holding the coarse readout fixed}
\begin{figure*}[t]
\global\QFourmeasureOverviewtrue
\centering\includegraphics[width=.96\textwidth]{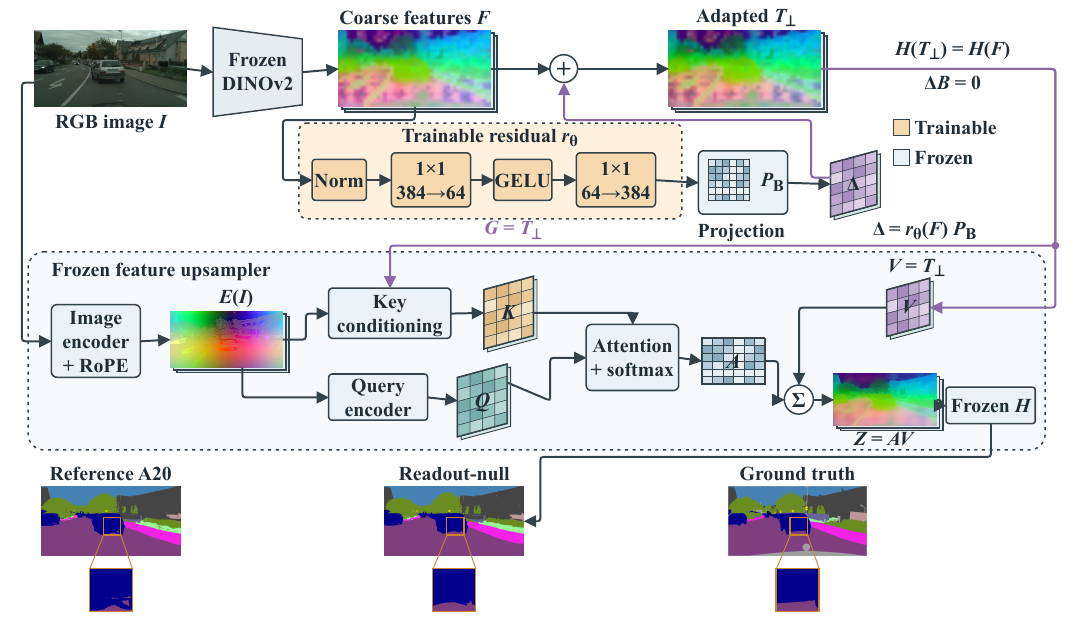}
\caption{Readout-null adaptation with an illustrative Cityscapes input. The adapted coarse field conditions the keys and supplies values, while the fixed projector preserves the affine coarse logits. Feature image fields are PCA views; grid glyphs are schematic.}
\label{fig:mechanism}
\end{figure*}

For the route-isolation experiment, let the frozen head be affine:
\begin{equation}
 H(Z)=ZB+\mathbf1 b^\top,\quad B\in\R^{D\times C}.
\end{equation}
Let $Q_B$ contain an orthonormal basis for the column space of $B$ and define
\begin{equation}
 \begin{aligned}
 P_B&=I_D-Q_BQ_B^\top,\\
 \Delta(F)&=R(F)P_B,\quad T_\perp(F)=F+\Delta(F).
 \end{aligned}
 \label{eq:null}
\end{equation}
Then $P_BB=0$, so $H(T_\perp(F))=H(F)$ at every coarse token. We use \emph{readout-null} to name this fixed-readout constraint; it preserves the classifier logits while leaving other feature information unrestricted. We obtain the projector from the frozen head and keep it fixed, adding no trainable parameters. Fig.~\ref{fig:mechanism} follows the constrained residual through the guidance and value paths. The diagrams write $r_\theta$ for the residual increment $R$. For visualization, $F$, $T_\perp(F)$ and $Z$ share a PCA basis and display scale; the image encoding $E(I)$ is projected separately.

For any fixed row-stochastic spatial matrix $A_0$, associativity gives
\begin{equation}
 H\!\left(A_0(F+\Delta)\right)=H(A_0F),
 \label{eq:fixed}
\end{equation}
because $A_0\Delta B=0$. Row normalization preserves the affine bias. We first set $A_0=A(I,F)$: null values alone cannot change the final logits under original guidance. Taking $A_0=A(I,F+\Delta)$ gives
\begin{equation}
 H\!\left(A(I,F+\Delta)(F+\Delta)\right)
 =H\!\left(A(I,F+\Delta)F\right).
 \label{eq:guide}
\end{equation}
A dense-logit change in this model is therefore carried by a change in the spatial weights. The accuracy experiment tests whether training makes these weight changes useful.

\subsection{Frozen-checkpoint exchanges and retraining}
We evaluate six forward conditions for each retained readout-null checkpoint. NativeA20 is the original reference forward; SplitA20 reproduces it with explicit guidance and value arguments. FixedGuideNull uses $A(I,F)$ with values $F+\Delta$. NativeFullNull and SplitFullNull use the adapted feature in both roles. NullGuideOnly keeps $A(I,F+\Delta)$ and restores values $F$. The native/split pairs check the explicit implementation, while the remaining exchanges test (\ref{eq:fixed}) and (\ref{eq:guide}). These are inference interventions on fixed parameters and entail no new fitting.

\begin{figure*}[t]
\centering\includegraphics[width=\textwidth]{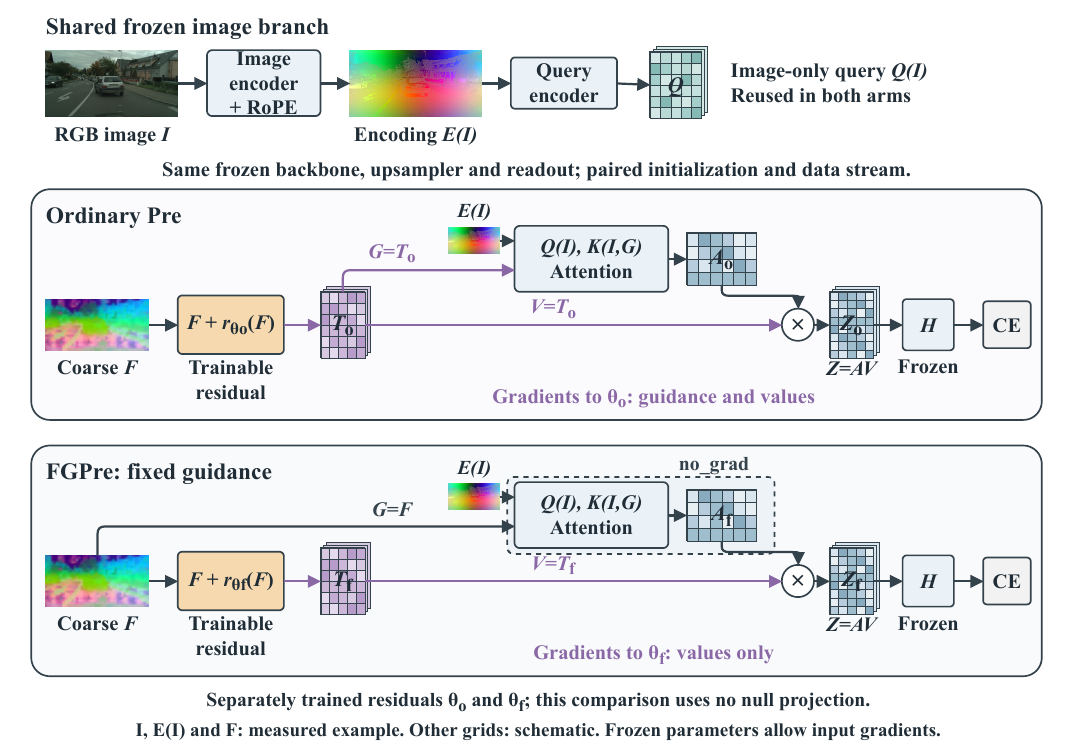}
\caption{Guidance and value paths in ordinary Pre and FGPre. Ordinary Pre uses $G=V=T_o(F)$; FGPre computes $A(I,F)$ under \texttt{no\_grad} and mixes differentiable $V=T_f(F)$. The paired VOC control uses separately trained residuals with identical initialization and data streams.}
\label{fig:guidance_control}
\end{figure*}

We name the separate retraining control \emph{fixed-guidance Pre} (FGPre). PreMLP, PostMLP, PreNull, and FGPre are labels for the strategies defined here. For FGPre, we compute keys and spatial attention weights from the original $F$ without gradients, while the trainable $T(F)$ supplies values. Its residual is unconstrained. The ordinary-Pre--FGPre comparison measures the joint effects of adaptive guidance on the forward function and optimization trajectory. Fig.~\ref{fig:guidance_control} distinguishes these two roles, with $T_o$ and $T_f$ denoting the respective fitted maps.

The ordinary residual class contains the projected residual class because $P_B$ can be absorbed into the output map. The two strategies therefore differ in their training constraints. We examine their learned solutions through attention and final-weight risk measurements.

\subsection{The second operator and the scope of the factorization}
The original AnyUp decoder also forms dense features by applying averaged spatial attention to unprojected values. Its image encoding and rotary position encoding supply image-only queries. Keys combine an image-derived field with an encoding of normalized coarse guidance features. We change which tensor supplies guidance and values through a split wrapper while preserving the pinned original decoder. We evaluate the original non-NATTEN decoder with 1,024 query positions per chunk. The AnyUp paper describes a window-attention variant. The affine-readout algebra applies to our selected decoder because it aggregates unprojected values. The full frozen-checkpoint exchanges above are measured for JAFAR; the AnyUp extension tests the two retrained comparisons under its own protocol.

\section{Experimental Setup}
\begin{table*}[t]
\centering\small
\caption{Experimental comparisons. Checkpoint exchanges and bilinear swaps use already trained models.}
\label{tab:families}
\begin{tabular}{llll}
\toprule
Setting & Comparison & Replication & Question \\
\midrule
JAFAR: VOC, Cityscapes & PreMLP vs. PostMLP & 3 seeds per dataset & Adapter placement \\
JAFAR: VOC, Cityscapes & PreNull vs. A20 & 3 seeds per dataset & Readout-null adaptation \\
JAFAR: VOC & ordinary Pre vs. FGPre & 3 paired seeds & Adaptive guidance \\
JAFAR: VOC & Six checkpoint-exchange conditions & 3 retained models & Guidance/value routes \\
JAFAR: Cityscapes & Bilinear swap; PreNull vs. FGPre & 3 training seeds & Operator and strategy controls \\
JAFAR: VOC & PreNull vs. ordinary Pre & 3 paired seeds & Attention and training risk \\
AnyUp: VOC, Cityscapes & PreNull vs. A20; ordinary Pre vs. FGPre & 3 seeds per dataset & Transfer across upsamplers \\
\bottomrule
\end{tabular}
\end{table*}

\subsection{Datasets and frozen models}
We use the PASCAL VOC segmentation split with 1,464 training and 1,449 validation images~\cite{everingham2010voc}, and Cityscapes with 2,975 training and 500 validation images and 19 official training classes~\cite{cordts2016cityscapes}. VOC images follow the reference short-edge resize and center crop to $448\times448$, yielding a $32\times32$ feature grid. Cityscapes uses a full-field $322\times644$ input and a $23\times46$ grid. We resize Cityscapes logits to the original $1024\times2048$ grid before taking the class argmax. We evaluate every model at its fixed final update.

DINOv2 ViT-S/14 supplies 384-dimensional features. We keep its weights and the reference JAFAR weights frozen. We run the new cohorts in FP32 with TF32 disabled. Images and labels use their frozen preprocessing rules; ignored labels are 255. Cityscapes tests transfer with both the dataset and input geometry changed. We use the official 19-class mapping for Cityscapes.

\subsection{Implementation details}
For each VOC seed, we fit a 20-epoch affine parent (A20) and a 30-epoch nonlinear parent (MLP142). We choose width 142 for the nonlinear parent to approximately match the A20-plus-residual parameter count. The placement branches PreMLP and PostMLP freeze that same MLP142 checkpoint. PreNull freezes A20. Every residual is a zero-output-initialized $384\!\to\!64\!\to\!384$ pointwise map with 49,600 parameters, trained for ten epochs. The paired guidance experiment trains ordinary Pre and FGPre from each of the same three A20 parents with shared residual initialization and streams. The independently trained Cityscapes parents use MLP width 141, followed by the same four residual strategies. Its FGPre is unconstrained and freezes A20, so PreNull--FGPre compares strategies that differ in both the null constraint and adaptive guidance.

We train with batch size two. A20 uses AdamW~\cite{loshchilov2019adamw} with peak learning rate $5\times10^{-4}$ and cosine decay to zero. Nonlinear parents and residuals use a peak of $10^{-3}$, one epoch of warmup, cosine decay to $10^{-4}$, weight decay $10^{-4}$, and gradient-norm clipping at one. A20 has no warmup or clipping. Frozen parameter gradients are disabled. We give paired branches the same image order and random horizontal flips. We label the three paired training seeds Seed 1, Seed 2, and Seed 3 within each cohort. All fits use AdamW with $(\beta_1,\beta_2)=(0.9,0.999)$ and $\epsilon=10^{-8}$.

\subsection{Evaluation and statistical analysis}
The primary outcome is pooled dataset mIoU, computed by summing per-image true-positive, prediction, and ground-truth counts before taking class ratios. A paired observation is the difference between two models trained under the same seed. Training seeds are the replication unit, $n=3$. For each dataset, we designate PreMLP--PostMLP and PreNull--A20 as the two co-primary contrasts. Each uses a two-sided 97.5\% Student-$t$ interval with two degrees of freedom, giving Bonferroni familywise 95\% coverage for that dataset's pair. The frozen success rule requires both mean differences to reach 0.20 pp, all three differences in each contrast to be positive, and both lower bounds to exceed zero. The familywise coverage applies separately to the VOC and Cityscapes pairs.

We report other training comparisons with nominal 95\% seed intervals and retain their secondary status. The $t$ intervals assume approximately normal paired seed differences; three seeds provide little information for checking that assumption. We therefore display all seed differences as well as the interval. Image and sequence bootstrap analyses quantify validation-sample uncertainty conditional on the fitted models. We use 2,000 paired bootstrap draws. A draw requires ground-truth support for every required class, and a conditional interval is reported only when at least 95\% of draws are defined.

\subsection{Tests with a second upsampler}
To test whether the guidance result depends on JAFAR's particular implementation, we replace its frozen operator with the original AnyUp paper checkpoint~\cite{wimmer2026anyup}. We retain the DINOv2 representation, datasets, preprocessing, FP32 arithmetic, final-update evaluation, and residual architecture. For each dataset and each new seed (Seeds 1--3), we fit a fresh 20-epoch AnyUp/A20 parent and three ten-epoch residuals: PreNull, ordinary Pre, and FGPre. The residuals share their own parent and initial residual state, image order, flips, and learning-rate schedule. This adds 24 fits and 24 final validation prediction sets. This extension tests the readout-null gain and adaptive-guidance utility.

This extension has one separate four-comparison family: PreNull--A20 and ordinary Pre--FGPre on each of the two datasets. Each contrast uses a two-sided 98.75\% seed-$t$ interval ($n=3$, $df=2$), giving Bonferroni familywise 95\% coverage across these four comparisons. The frozen criterion retains a mean of at least 0.20 pp, three positive seed differences, and a positive lower bound for every contrast. Only the new AnyUp cohort enters this family. Conditional image and sequence resampling follows the same class-availability rule as before.

\subsection{Measurements of learned guidance}
The readout identities motivate a second question: how do the learned spatial weights change? We therefore measure JAFAR attention from the retained ordinary-Pre and PreNull models on all 1,449 VOC validation images. Original-feature guidance is shared as a reference, and uniform spatial weighting supplies a ground-truth-only reference. We record attention mass by key--query distance, same-class keys, cross-class keys, and void keys. Distances run from each query pixel center to a key-patch center and are divided by 14 image pixels. Ground-truth class fractions within each coarse key cell account for mixed cells; void fractions remain in the mass partition. Queries with ignored labels are excluded.

The primary aggregation gives each eligible image equal weight. We also retain query-weighted statistics, eligible-image counts, and paired per-image values. Conditional same-class fractions are formed from the corresponding pooled same- and cross-class masses; a zero denominator is recorded as unavailable. These summaries describe spatial associations in the trained models. Their relation to accuracy is considered alongside the readout interventions.

We define \emph{canonical final-weight training risk} as evaluation of the final JAFAR-system weights on all 1,464 canonical training images with no random flip. Cross-entropy is weighted by the total number of valid pixels; mIoU pools class counts before taking ratios. We compare each ordinary-Pre/PreNull pair trained from the same A20 parent and residual initialization. This deterministic evaluation measures final-weight fit on unflipped inputs; the online objective uses random flips. We compare training fit with validation performance to examine how the two learned solutions differ.


\section{Results}
\begin{table}[t]
\centering
\caption{JAFAR strategies and task-head-plus-adapter parameter counts. ``Parent'' is trained first and then frozen; the trainable residual has 49,600 parameters in every residual branch. VOC/City counts reflect 21/19 output classes.}
\label{tab:design}
\small
\begin{tabular}{llrr}
\toprule
Strategy & Frozen parent & VOC total & City total \\
\midrule
A20 & None (parent) & 8,085 & 7,315 \\
MLP142/141 & None (parent) & 57,673 & 56,983 \\
PreMLP & MLP142/141 & 107,273 & 106,583 \\
PostMLP & Same MLP & 107,273 & 106,583 \\
ordinary Pre & A20 & 57,685 & Not fitted \\
PreNull & A20 & 57,685 & 56,915 \\
FGPre & A20 & 57,685 & 56,915 \\
\bottomrule
\end{tabular}
\end{table}

\begin{table*}[t]
\centering
\caption{Final pooled mIoU (\%) for the new training cohorts. Each column is one paired training seed; $\pm$ denotes sample SD over three seeds. VOC guidance controls use the same A20 parents but fresh paired residual training. Cityscapes includes a zero-fit bilinear swap.}
\label{tab:endpoints}
\small
\begin{tabular}{llrrrr}
\toprule
Setting & Strategy & Seed 1 & Seed 2 & Seed 3 & Mean $\pm$ SD \\
\midrule
VOC confirmation & A20 & 84.300 & 84.339 & 84.390 & $84.343\pm0.045$ \\
VOC confirmation & MLP142 & 84.562 & 85.060 & 84.914 & $84.845\pm0.256$ \\
VOC confirmation & PreMLP & 85.355 & 85.896 & 85.612 & $85.621\pm0.270$ \\
VOC confirmation & PostMLP & 84.571 & 85.034 & 84.839 & $84.814\pm0.232$ \\
VOC confirmation & PreNull & 85.712 & 85.736 & 85.787 & $85.745\pm0.039$ \\
\midrule
VOC guidance & ordinary Pre & 85.411 & 85.487 & 85.492 & $85.463\pm0.045$ \\
VOC guidance & FGPre & 84.446 & 84.583 & 84.556 & $84.528\pm0.072$ \\
\midrule
Cityscapes & A20 & 61.971 & 61.973 & 62.012 & $61.986\pm0.023$ \\
Cityscapes & MLP141 & 62.409 & 62.671 & 62.462 & $62.514\pm0.139$ \\
Cityscapes & PreMLP & 63.834 & 64.075 & 64.117 & $64.008\pm0.153$ \\
Cityscapes & PostMLP & 62.502 & 62.751 & 62.534 & $62.596\pm0.135$ \\
Cityscapes & PreNull & 63.923 & 63.956 & 64.040 & $63.973\pm0.060$ \\
Cityscapes & FGPre & 61.957 & 61.969 & 62.021 & $61.983\pm0.034$ \\
Cityscapes & Bilinear (A20) & 55.232 & 55.217 & 55.047 & $55.165\pm0.103$ \\
\bottomrule
\end{tabular}
\end{table*}

\subsection{Locating the benefit: before versus after upsampling}
\begin{figure*}[t]
\centering
\includegraphics[width=\textwidth]{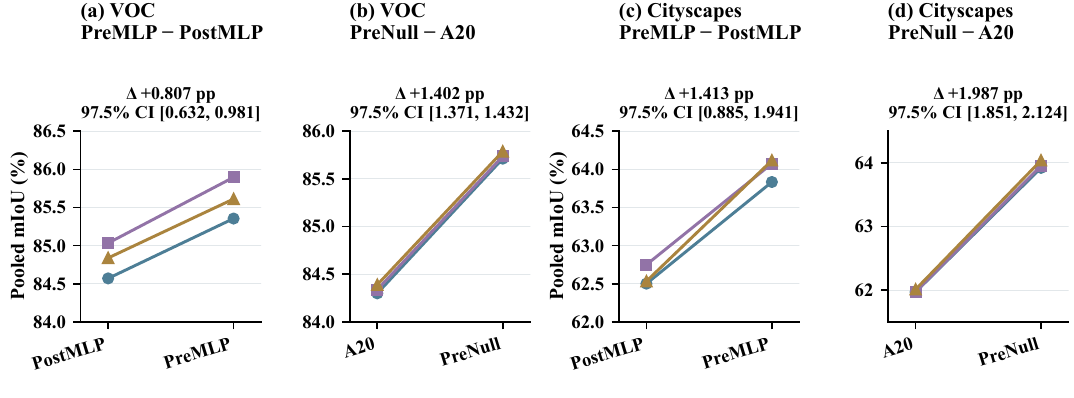}
\caption{Paired JAFAR endpoints across three training seeds. Text reports mean differences and 97.5\% seed-$t$ intervals for the two primary comparisons within each dataset.}
\label{fig:seeds}
\end{figure*}

Table~\ref{tab:endpoints} reports the complete seed endpoints. In the new VOC cohort, PreMLP reaches $85.621\%$ mean mIoU, compared with $84.814\%$ for PostMLP. The paired advantage is $0.807$ pp with a 97.5\% seed interval of $[0.632,0.981]$ pp. All three differences are positive. Both adapters freeze the same MLP142, so the placement advantage persists under this stronger nonlinear parent. PreMLP also exceeds that parent alone by $0.776$ pp, with a nominal 95\% interval of $[0.599,0.952]$ pp. Fig.~\ref{fig:seeds} shows the paired endpoints for every seed.

PostMLP changes the mean little relative to MLP142: the paired difference is $-0.031$ pp. In a retrospective equivalence assessment, we find that the 90\% seed interval, $[-0.102,0.040]$ pp, lies within $\pm0.20$ pp. A separate conditional paired-image interval is $[-0.078,0.017]$ pp. We reused the margin after observing the results. The seed and conditional image intervals describe different sources of uncertainty. Together with the positive PreMLP contrast, they place the useful adaptation before upsampling for this parent and training recipe.

On Cityscapes, PreMLP exceeds PostMLP by $1.413$ pp, with a 97.5\% interval of $[0.885,1.941]$ pp. PreMLP exceeds its MLP141 parent by $1.495$ pp with nominal 95\% interval $[1.149,1.840]$ pp.

\subsection{Testing adaptation with the coarse readout fixed}
We obtain $85.745\%$ mean VOC mIoU with PreNull, an improvement of $1.402$ pp over A20. Its 97.5\% seed interval is $[1.371,1.432]$ pp; each paired difference is positive. This branch has 57,685 total head-plus-residual parameters, close to the 57,673 parameters of MLP142, yet its defining constraint keeps the original A20 coarse logits unchanged in exact arithmetic. The dense gain is obtained with the coarse linear classifier output held fixed.

On Cityscapes, PreNull exceeds A20 by $1.987$ pp, with interval $[1.851,2.124]$ pp. All three paired differences are positive. Both co-primary criteria pass for each dataset.

\begin{table}[t]
\centering
\caption{VOC frozen-checkpoint exchanges: mean mIoU (\%) over the same three readout-null checkpoints. Six decimal places show numerical differences. All conditions are zero-fit forward interventions.}
\label{tab:null}
\small
\begin{tabular}{lccr}
\toprule
Condition & Guidance & Values & Mean \\
\midrule
NativeA20 & Original & Original & 84.343330 \\
SplitA20 & Original & Original & 84.343330 \\
FixedGuideNull & Original & Null-adapted & 84.343323 \\
NativeFullNull & Adapted & Null-adapted & 85.744981 \\
SplitFullNull & Adapted & Null-adapted & 85.744981 \\
NullGuideOnly & Adapted & Original & 85.744980 \\
\bottomrule
\end{tabular}
\end{table}

We check the implementation of the route argument through retained-checkpoint interventions (Table~\ref{tab:null}). We recover the same mIoU from native and split original inference. FixedGuideNull returns to the original score, whereas NullGuideOnly retains the full adapted score. The mean numerical differences relative to the corresponding split forward are approximately $-7.0\times10^{-6}$ pp and $-0.9\times10^{-6}$ pp. Small mask differences can remain near argmax ties under finite-precision computation. All 18 prediction sets were independently reconstructed from retained maps and counts. The direction and scale agree with (\ref{eq:fixed})--(\ref{eq:guide}), identifying guidance as the route through which this constrained model changes its dense readout.

\subsection{Retraining with and without adaptive guidance}
In the separate paired VOC guidance cohort, we obtain $85.463\%$ with ordinary Pre, while FGPre reaches $84.528\%$. Their mean difference is $0.935$ pp, with nominal 95\% seed interval $[0.861,1.010]$ pp; all three differences are positive. Adaptive feature-conditioned guidance retains its advantage when both strategies are trained from the same residual initial states and frozen A20 parents. This retraining result complements the fixed-checkpoint exchanges by measuring the performance attained with adaptive guidance during optimization.

\begin{table}[t]
\centering\small
\caption{Measured JAFAR VOC guidance-training cost: 7,320 updates per branch on the same RTX 4090 cloud instance (47.4 GiB reported device memory). Time is training-loop wall time (mean $\pm$ sample SD); memory is mean peak PyTorch allocation.}
\label{tab:cost}
\begin{tabular}{lrr}
\toprule
Strategy & Time (s) & Peak allocation (GiB) \\
\midrule
ordinary Pre & $1640.1\pm1.2$ & 24.80 \\
FGPre & $1071.1\pm0.6$ & 13.30 \\
\bottomrule
\end{tabular}
\end{table}

We also measure the resource cost of adaptive guidance. In the matched-update VOC comparison, we find that ordinary Pre takes 1.531 times the training-loop wall time and 1.864 times the peak PyTorch-allocated memory of FGPre (Table~\ref{tab:cost}). The timed loop includes epoch checkpoints and control checks but excludes feature preparation, final validation, and export. These implementation-specific measurements place the accuracy gain at matched updates and residual capacity alongside its additional computational cost.

\subsection{Dataset consistency and remaining strategy differences}
On Cityscapes, we obtain $55.165\%$ mean mIoU after swapping in bilinear interpolation, compared with $61.986\%$ for the A20/JAFAR system. This zero-fit swap measures the effect of replacing the upsampler under the same trained A20 head.

Cityscapes PreNull exceeds FGPre by $1.990$ pp, with nominal 95\% interval $[1.924,2.056]$ pp. Because this contrast changes the null constraint as well as adaptive guidance, its conclusion is about the complete trained strategies. The AnyUp extension below supplies the matched ordinary-Pre--FGPre comparison on Cityscapes.

\begin{figure*}[t]
\centering\includegraphics[width=\textwidth]{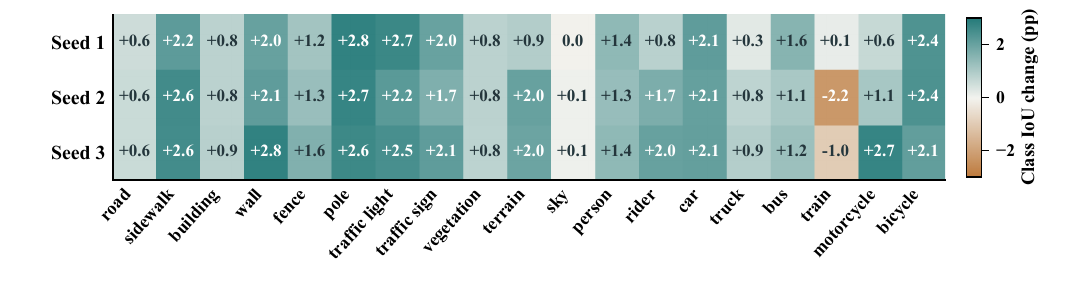}
\caption{Cityscapes class-IoU effects for PreMLP minus PostMLP, computed from pooled class counts. All three seeds and 19 official classes are shown on one centered colour scale.}
\label{fig:class_heterogeneity}
\end{figure*}

The class effects in Fig.~\ref{fig:class_heterogeneity} show that an overall mIoU gain is not uniform improvement for every class and seed.

\subsection{Repeating the mechanism tests with AnyUp}

\begin{table*}[t]
\centering\small
\caption{AnyUp pooled mIoU (\%): all training seeds and mean $\pm$ sample standard deviation. Each residual freezes its own seed-matched A20 parent.}
\label{tab:anyup}
\begin{tabular}{llrrrr}
\toprule
Setting & Seed & A20 & PreNull & Ordinary Pre & FGPre \\
\midrule
VOC & 1 & 84.145 & 85.530 & 84.912 & 84.417 \\
 & 2 & 84.403 & 85.773 & 85.134 & 84.618 \\
 & 3 & 84.340 & 85.720 & 85.004 & 84.507 \\
 & Mean $\pm$ SD & $84.296\pm0.135$ & $85.674\pm0.127$ & $85.017\pm0.112$ & $84.514\pm0.101$ \\
\midrule
Cityscapes & 1 & 61.605 & 63.057 & 62.777 & 61.600 \\
 & 2 & 61.565 & 63.001 & 62.804 & 61.576 \\
 & 3 & 61.618 & 63.037 & 62.799 & 61.613 \\
 & Mean $\pm$ SD & $61.596\pm0.028$ & $63.032\pm0.028$ & $62.794\pm0.015$ & $61.596\pm0.019$ \\
\bottomrule
\end{tabular}
\end{table*}

\begin{figure*}[t]

\centering\includegraphics[width=\textwidth]{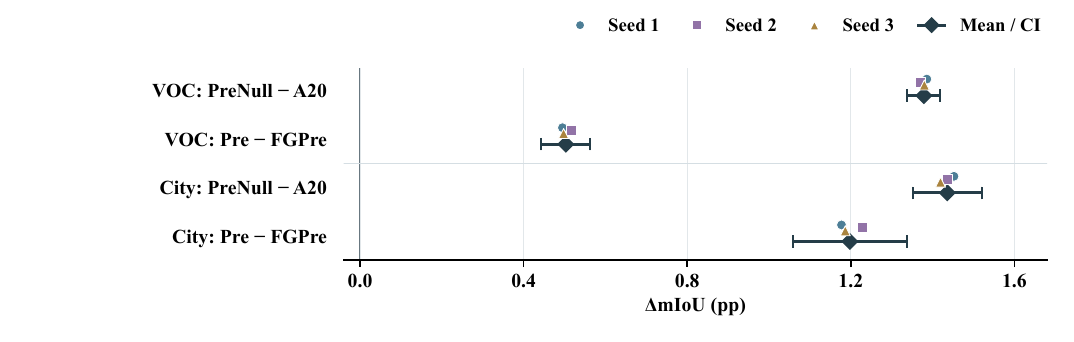}

\caption{Paired AnyUp improvements. Small markers show three training seeds; diamonds and bars show means and 98.75\% seed-$t$ intervals, adjusted across the four contrasts.}

\label{fig:anyup}

\end{figure*}

We replace JAFAR with AnyUp to test whether the constrained adaptation result and guidance utility depend on the first operator. Each AnyUp parent is fitted anew, so its baseline is the matching AnyUp/A20 system. Table~\ref{tab:anyup} gives the endpoints and Fig.~\ref{fig:anyup} shows every paired seed effect.

On VOC, we find that PreNull exceeds its A20 parent by $1.379$ pp, with a 98.75\% seed interval of $[1.338,1.419]$ pp. Ordinary Pre exceeds FGPre by $0.503$ pp, with interval $[0.443,0.563]$ pp.

On Cityscapes, we find that PreNull exceeds its A20 parent by $1.436$ pp, with a 98.75\% seed interval of $[1.352,1.520]$ pp. Ordinary Pre exceeds FGPre by $1.197$ pp, with interval $[1.058,1.337]$ pp.

All three seed differences are positive in every contrast, and all four comparisons pass the fixed mean, direction, and adjusted-interval criteria. This family is separate from the earlier JAFAR comparisons. AnyUp supports the two retrained comparisons under a second operator with unprojected attention aggregation. JAFAR supplies the nonlinear-parent placement and full checkpoint-exchange evidence.

\section{Analysis and Limitations}
\subsection{What the intervention establishes}
The placement experiment establishes a useful difference between compositions with matched residual capacity and a common strong parent. The null experiment is more specific: within a frozen affine readout and shared linear value aggregation, it constructs a useful residual whose direct coarse-readout effect is zero. The forward exchanges verify the implementation-level consequences of that restriction. Fixed-guidance retraining shows that disabling the adaptive route changes attainable performance under the chosen recipe. The AnyUp extension supports the null and retrained-guidance results with a second operator on both datasets. Together, these tests link the adaptation gain to feature-conditioned spatial guidance in the tested systems.

\subsection{How learned spatial guidance changes}
In an exploratory comparison sharing A20 parents and initial states, we find that PreNull exceeds ordinary Pre by 0.282 pp on average, with all three differences positive. This contrast motivates examining the two strategies' final behavior; it is outside the co-primary family.

\begin{figure*}[t]
\centering\includegraphics[width=\textwidth]{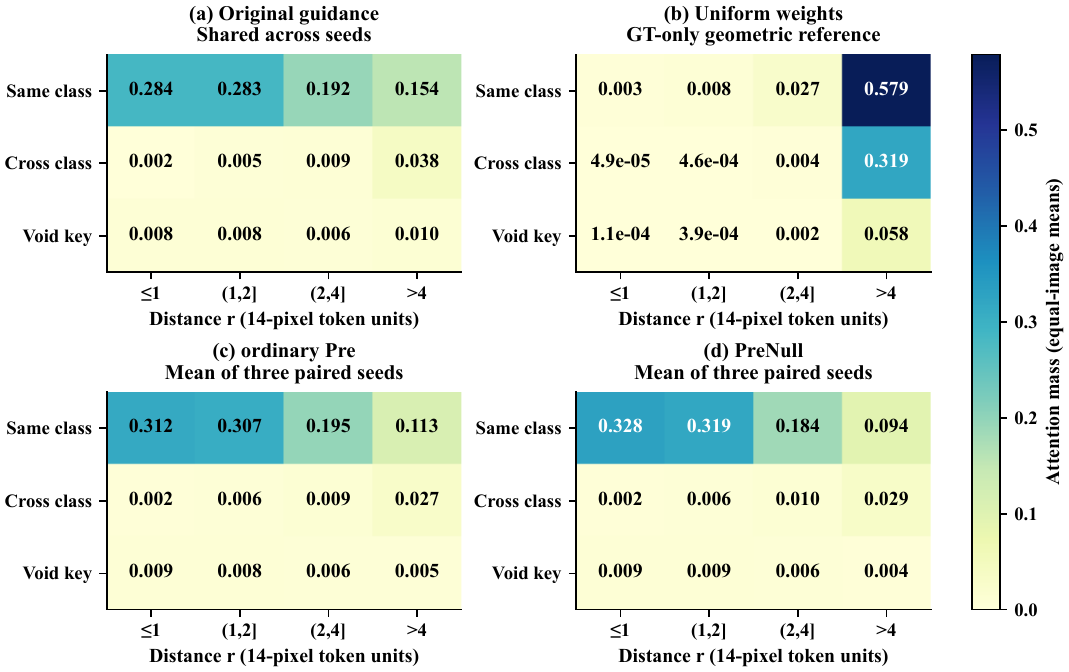}
\caption{Attention mass by distance and ground-truth key class. Cells are equal-image means; adapted panels average three seeds, while each shared reference appears once. Distance is in 14-pixel units. All panels use the same color scale.}
\label{fig:attention}
\end{figure*}

In Fig.~\ref{fig:attention}, PreNull outperforms ordinary Pre while assigning less total attention mass to the query class. Across the three paired seeds, PreNull assigns less same-class mass, more cross-class mass, and less void mass when valid queries are averaged within each image and images are weighted equally. Its expected query-to-key distance is also smaller. Relative to the original guidance, both adaptations increase same-class mass and shorten expected distance. Their difference is in the spatial distribution of that mass. Compared with ordinary Pre, we find that PreNull increases same-class mass in the two bands within two token units and decreases it in both more distant bands in every seed. The near/far redistribution describes the learned checkpoints. The readout intervention supplies the separate evidence that guidance can improve prediction while preserving the coarse affine logits.

\subsection{Training fit and validation performance}
\begin{figure}[t]
\centering\includegraphics[width=\columnwidth]{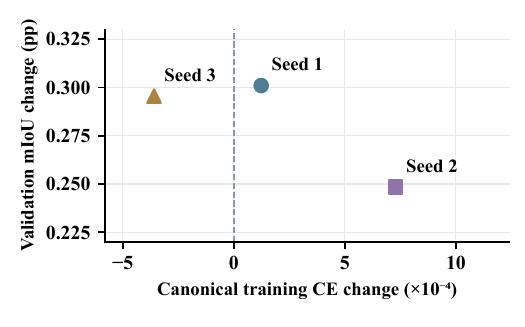}
\caption{Validation gains versus canonical training cross-entropy changes (PreNull minus ordinary Pre). Each point shows one paired training-seed contrast.}
\label{fig:risk_tradeoff}
\end{figure}

We compare the training and validation changes in Fig.~\ref{fig:risk_tradeoff}. On the 1,464 training images under the canonical deterministic transform, the PreNull-minus-Pre cross-entropy differences are $+0.000123$, $+0.000729$, and $-0.000359$ for Seeds 1, 2, and 3, respectively. Training mIoU increases by $0.0723$, $0.0185$, and $0.0693$ percentage points. On the 1,449 validation images, we find lower PreNull cross-entropy in all three seeds and gains of $0.3010$, $0.2484$, and $0.2955$ mIoU points. The validation gain is consistent across seeds, while the canonical training cross-entropy changes have mixed signs. The final-weight measurements describe this difference; explaining its optimization history requires additional evidence.

\subsection{Spatial examples across operators}
\begin{figure*}[t]
\centering
\includegraphics[width=\textwidth]{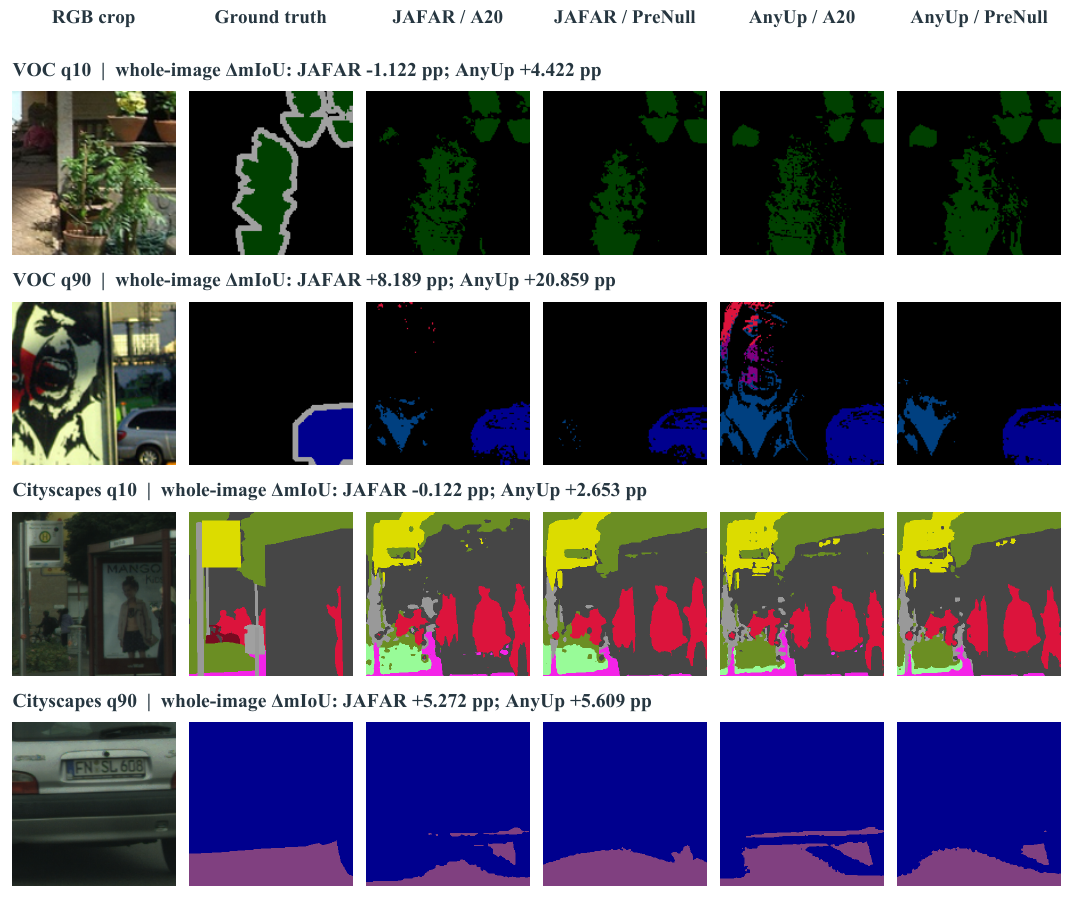}
\caption{Real outputs of both frozen operators on the original JAFAR-selected q10/q90 cases and crops. Row annotations are whole-image PreNull--A20 differences; q10/q90 refer only to JAFAR selection. The first seed of each operator is used, with a shared class palette within each dataset.}
\label{fig:cases}
\end{figure*}

We retain the original low- and high-quantile JAFAR cases and add AnyUp predictions at exactly the same image IDs and regions in Fig.~\ref{fig:cases}. The case selection is fixed across operators. The original selection used retained whole-image count arrays before mask inspection. We sort single-image mIoU differences and then image IDs, taking zero-based rank \texttt{round(q*(n-1))} for $q=0.10$ and $0.90$. The crop maximizes valid prediction disagreement on a stride-16 grid, using 144 pixels on VOC and 256 native pixels on Cityscapes; coordinate ties choose the smallest top and then left position. The last row revisits the car in Fig.~\ref{fig:problem}: both operators change its dense boundary despite the fixed coarse readout. At the VOC q10 case, the whole-image effect is negative for JAFAR and positive for AnyUp, illustrating heterogeneity across operators. We use the first seed within each operator and dataset cohort. Seed labels are local to their cohort, and row annotations give whole-image effects. The examples illustrate spatial behavior; quantitative operator-specific conclusions use the complete validation sets. The reproduction package retains median examples, correction/regression maps, selection rules, and exact identities.

\subsection{Limitations and scope}
The algebra requires unprojected value aggregation and a fixed affine head. We test these conditions with one DINOv2 representation and two frozen upsamplers. The statistical unit is the paired training seed; image and cluster resampling quantify conditional sample uncertainty. The evidence covers VOC crops and Cityscapes full fields under the stated protocols. Architectures with learned value projections or nonlinear readouts fall outside the algebra used here. Both public validation sets were examined during study development; the new seeds quantify training variability on those datasets. Development cohorts and the single-seed screen remain separate from the primary training replicates.

The validation images form 235 city--sequence clusters. Of 2,000 paired cluster-bootstrap draws, 1,728 had defined values for every required class, giving $86.4\%$ availability. This falls below the frozen 95\% availability threshold, so the conditional cluster interval is unavailable. The primary decisions continue to use the training-seed intervals defined in the protocol.

\section{Conclusion}
We find that a task adapter can improve frozen feature upsampling by changing feature-conditioned spatial guidance. Matched-parent experiments show a pre-upsampling advantage on VOC and Cityscapes under their respective input geometries. Readout-null adaptation preserves coarse affine logits yet improves dense prediction; checkpoint exchanges verify the resulting readout identities. Retrained fixed-guidance controls support the usefulness of adapting spatial weights under the tested recipes. The independent AnyUp extension tests the same null and guidance comparisons in a second operator on both datasets. Together, the results establish adaptive spatial guidance as a useful task-adaptation route in the two tested attention systems.

\section*{Acknowledgment}
The author used GPT (OpenAI) for language editing throughout the manuscript. The author takes full responsibility for the content of this article.
\bibliographystyle{IEEEtran}
\bibliography{references}

@IEEEtranBSTCTL{IEEEcontrol,
  CTLuse_forced_etal = {yes},
  CTLmax_names_forced_etal = {6},
  CTLnames_show_etal = {1}
}

@inproceedings{caron2021dino,
  author = {Caron, Mathilde and Touvron, Hugo and Misra, Ishan and J{\'e}gou, Herv{\'e} and Mairal, Julien and Bojanowski, Piotr and Joulin, Armand},
  title = {{Emerging properties in self-supervised vision transformers}},
  year = {2021},
  booktitle = {Proc. IEEE/CVF Int. Conf. Comput. Vis. (ICCV)},
  pages = {9630--9640},
  doi = {10.1109/iccv48922.2021.00951},
  note = {doi: \href{https://doi.org/10.1109/iccv48922.2021.00951}{\nolinkurl{10.1109/iccv48922.2021.00951}}}
}

@inproceedings{chambon2025naf,
  author = {Chambon, Loick and Couairon, Paul and Zablocki, Eloi and Boulch, Alexandre and Thome, Nicolas and Cord, Matthieu},
  title = {{NAF: Zero-shot feature upsampling via neighborhood attention filtering}},
  year = {2026},
  booktitle = {Proc. IEEE/CVF Conf. Comput. Vis. Pattern Recognit. (CVPR)},
  url = {https://valeoai.github.io/publications/naf/},
  pages = {26604--26613}
}

@inproceedings{chen2018deeplabv3p,
  author = {Chen, Liang-Chieh and Zhu, Yukun and Papandreou, George and Schroff, Florian and Adam, Hartwig},
  title = {{Encoder-decoder with atrous separable convolution for semantic image segmentation}},
  year = {2018},
  booktitle = {Proc. Eur. Conf. Comput. Vis. (ECCV)},
  pages = {833--851},
  doi = {10.1007/978-3-030-01234-2_49},
  note = {doi: \href{https://doi.org/10.1007/978-3-030-01234-2_49}{\nolinkurl{10.1007/978-3-030-01234-2_49}}}
}

@inproceedings{chen2022adaptformer,
  author = {Chen, Shoufa and Ge, Chongjian and Tong, Zhan and Wang, Jiangliu and Song, Yibing and Wang, Jue and Luo, Ping},
  title = {{AdaptFormer: Adapting vision transformers for scalable visual recognition}},
  year = {2022},
  booktitle = {Adv. Neural Inf. Process. Syst. (NeurIPS)},
  pages = {16664--16678},
  volume = {35},
  doi = {10.52202/068431-1212},
  note = {doi: \href{https://doi.org/10.52202/068431-1212}{\nolinkurl{10.52202/068431-1212}}}
}

@article{chen2026clues,
  author = {Chen, Xiaoyu and Zou, Qi and Wang, Jing and Guo, Jianyong and Wu, Xinyi and Wang, Song and Wu, Zhenyao},
  title = {{Injecting task-specific clues: Efficient adaptation of vision foundation models for domain-generalized semantic segmentation}},
  year = {2026},
  doi = {10.1109/tcsvt.2026.3689391},
  journal = {IEEE Trans. Circuits Syst. Video Technol.},
  volume = {36},
  number = {9},
  pages = {13011--13022},
  note = {doi: \href{https://doi.org/10.1109/tcsvt.2026.3689391}{\nolinkurl{10.1109/tcsvt.2026.3689391}}}
}

@inproceedings{cheng2022mask2former,
  author = {Cheng, Bowen and Misra, Ishan and Schwing, Alexander G. and Kirillov, Alexander and Girdhar, Rohit},
  title = {{Masked-attention mask transformer for universal image segmentation}},
  year = {2022},
  booktitle = {Proc. IEEE/CVF Conf. Comput. Vis. Pattern Recognit. (CVPR)},
  pages = {1280--1289},
  doi = {10.1109/cvpr52688.2022.00135},
  note = {doi: \href{https://doi.org/10.1109/cvpr52688.2022.00135}{\nolinkurl{10.1109/cvpr52688.2022.00135}}}
}

@inproceedings{cordts2016cityscapes,
  author = {Cordts, Marius and Omran, Mohamed and Ramos, Sebastian and Rehfeld, Timo and Enzweiler, Markus and Benenson, Rodrigo and Franke, Uwe and Roth, Stefan and Schiele, Bernt},
  title = {{The Cityscapes dataset for semantic urban scene understanding}},
  year = {2016},
  booktitle = {Proc. IEEE Conf. Comput. Vis. Pattern Recognit. (CVPR)},
  pages = {3213--3223},
  doi = {10.1109/cvpr.2016.350},
  note = {doi: \href{https://doi.org/10.1109/cvpr.2016.350}{\nolinkurl{10.1109/cvpr.2016.350}}}
}

@inproceedings{couairon2025jafar,
  author = {Couairon, Paul and Chambon, Loick and Serrano, Louis and Haugeard, Jean-Emmanuel and Cord, Matthieu and Thome, Nicolas},
  title = {{JAFAR: Jack up any feature at any resolution}},
  year = {2025},
  booktitle = {Adv. Neural Inf. Process. Syst. (NeurIPS)},
  doi = {10.52202/085713-3316},
  volume = {38},
  pages = {99265--99292},
  note = {doi: \href{https://doi.org/10.52202/085713-3316}{\nolinkurl{10.52202/085713-3316}}}
}

@inproceedings{dosovitskiy2021vit,
  author = {Dosovitskiy, Alexey and Beyer, Lucas and Kolesnikov, Alexander and Weissenborn, Dirk and Zhai, Xiaohua and Unterthiner, Thomas and Dehghani, Mostafa and Minderer, Matthias and Heigold, Georg and Gelly, Sylvain and Uszkoreit, Jakob and Houlsby, Neil},
  title = {{An image is worth 16x16 words: Transformers for image recognition at scale}},
  year = {2021},
  booktitle = {Proc. Int. Conf. Learn. Represent. (ICLR)},
  url = {https://openreview.net/forum?id=YicbFdNTTy}
}

@article{everingham2010voc,
  author = {Everingham, Mark and Van Gool, Luc and Williams, Christopher K. I. and Winn, John and Zisserman, Andrew},
  title = {{The Pascal Visual Object Classes (VOC) challenge}},
  year = {2010},
  doi = {10.1007/s11263-009-0275-4},
  journal = {Int. J. Comput. Vis.},
  volume = {88},
  number = {2},
  pages = {303--338},
  note = {doi: \href{https://doi.org/10.1007/s11263-009-0275-4}{\nolinkurl{10.1007/s11263-009-0275-4}}}
}

@inproceedings{fu2024featup,
  author = {Fu, Stephanie and Hamilton, Mark and Brandt, Laura E. and Feldmann, Axel and Zhang, Zhoutong and Freeman, William T.},
  title = {{FeatUp: A model-agnostic framework for features at any resolution}},
  year = {2024},
  booktitle = {Proc. Int. Conf. Learn. Represent. (ICLR)},
  pages = {45324--45350},
  url = {https://proceedings.iclr.cc/paper_files/paper/2024/hash/c5601d99ed028448f29d1dae2e4a926d-Abstract-Conference.html}
}

@article{he2010guided,
  author = {He, Kaiming and Sun, Jian and Tang, Xiaoou},
  title = {{Guided image filtering}},
  year = {2013},
  doi = {10.1109/tpami.2012.213},
  pages = {1397--1409},
  journal = {IEEE Trans. Pattern Anal. Mach. Intell.},
  volume = {35},
  number = {6},
  note = {doi: \href{https://doi.org/10.1109/tpami.2012.213}{\nolinkurl{10.1109/tpami.2012.213}}}
}

@inproceedings{he2022mae,
  author = {He, Kaiming and Chen, Xinlei and Xie, Saining and Li, Yanghao and Dollár, Piotr and Girshick, Ross},
  title = {{Masked autoencoders are scalable vision learners}},
  year = {2022},
  booktitle = {Proc. IEEE/CVF Conf. Comput. Vis. Pattern Recognit. (CVPR)},
  pages = {15979--15988},
  doi = {10.1109/cvpr52688.2022.01553},
  note = {doi: \href{https://doi.org/10.1109/cvpr52688.2022.01553}{\nolinkurl{10.1109/cvpr52688.2022.01553}}}
}

@inproceedings{hu2022lora,
  author = {Hu, Edward J. and Shen, Yelong and Wallis, Phillip and Allen-Zhu, Zeyuan and Li, Yuanzhi and Wang, Shean and Wang, Lu and Chen, Weizhu},
  title = {{LoRA: Low-rank adaptation of large language models}},
  year = {2022},
  booktitle = {Proc. Int. Conf. Learn. Represent. (ICLR)},
  url = {https://openreview.net/forum?id=nZeVKeeFYf9}
}

@inproceedings{huang2025loftup,
  author = {Huang, Haiwen and Chen, Anpei and Havrylov, Volodymyr and Geiger, Andreas and Zhang, Dan},
  title = {{LoftUp: Learning a coordinate-based feature upsampler for vision foundation models}},
  year = {2025},
  booktitle = {Proc. IEEE/CVF Int. Conf. Comput. Vis. (ICCV)},
  pages = {9913--9923},
  doi = {10.1109/iccv51701.2025.00924},
  note = {doi: \href{https://doi.org/10.1109/iccv51701.2025.00924}{\nolinkurl{10.1109/iccv51701.2025.00924}}}
}

@inproceedings{jie2022convpass,
  author = {Jie, Shibo and Deng, Zhi-Hong and Chen, Shixuan and Jin, Zhijuan},
  title = {{Convolutional bypasses are better vision transformer adapters}},
  year = {2024},
  doi = {10.3233/faia240489},
  booktitle = {Proc. Eur. Conf. Artif. Intell. (ECAI)},
  pages = {202--209},
  note = {doi: \href{https://doi.org/10.3233/faia240489}{\nolinkurl{10.3233/faia240489}}}
}

@article{kopf2007jbu,
  author = {Kopf, Johannes and Cohen, Michael F. and Lischinski, Dani and Uyttendaele, Matt},
  title = {{Joint bilateral upsampling}},
  year = {2007},
  doi = {10.1145/1276377.1276497},
  journal = {ACM Trans. Graph.},
  volume = {26},
  number = {3},
  pages = {96},
  note = {doi: \href{https://doi.org/10.1145/1276377.1276497}{\nolinkurl{10.1145/1276377.1276497}}}
}

@inproceedings{long2015fcn,
  author = {Long, Jonathan and Shelhamer, Evan and Darrell, Trevor},
  title = {{Fully convolutional networks for semantic segmentation}},
  year = {2015},
  booktitle = {Proc. IEEE Conf. Comput. Vis. Pattern Recognit. (CVPR)},
  pages = {3431--3440},
  doi = {10.1109/cvpr.2015.7298965},
  note = {doi: \href{https://doi.org/10.1109/cvpr.2015.7298965}{\nolinkurl{10.1109/cvpr.2015.7298965}}}
}

@inproceedings{loshchilov2019adamw,
  author = {Loshchilov, Ilya and Hutter, Frank},
  title = {{Decoupled weight decay regularization}},
  year = {2019},
  booktitle = {Proc. Int. Conf. Learn. Represent. (ICLR)},
  url = {https://openreview.net/forum?id=Bkg6RiCqY7}
}

@article{oquab2024dinov2,
  author = {Oquab, Maxime and Darcet, Timoth{\'e}e and Moutakanni, Th{\'e}o and Vo, Huy V. and Szafraniec, Marc and Khalidov, Vasil and Fernandez, Pierre and Haziza, Daniel and Massa, Francisco and El-Nouby, Alaaeldin and Assran, Mido and Ballas, Nicolas and Galuba, Wojciech and Howes, Russell and Huang, Po-Yao and Li, Shang-Wen and Misra, Ishan and Rabbat, Michael and Sharma, Vasu and Synnaeve, Gabriel and Xu, Hu and Jegou, Herv{\'e} and Mairal, Julien and Labatut, Patrick and Joulin, Armand and Bojanowski, Piotr},
  title = {{DINOv2: Learning robust visual features without supervision}},
  year = {2024},
  journal = {Transactions on Machine Learning Research},
  url = {https://openreview.net/forum?id=a68SUt6zFt}
}

@inproceedings{ranzinger2025featsharp,
  author = {Ranzinger, Mike and Heinrich, Greg and Molchanov, Pavlo and Catanzaro, Bryan and Tao, Andrew},
  title = {{FeatSharp: Your vision model features, sharper}},
  year = {2025},
  booktitle = {Proc. Int. Conf. Mach. Learn. (ICML)},
  volume = {267},
  pages = {51156--51182},
  url = {https://proceedings.mlr.press/v267/ranzinger25a.html}
}

@inproceedings{seo2025upa,
  author = {Seo, Minseok and Hamilton, Mark and Kim, Changick},
  title = {{Upsample Anything: A simple and hard to beat baseline for feature upsampling}},
  year = {2026},
  url = {https://openaccess.thecvf.com/CVPR2026?day=2026-06-07},
  booktitle = {Proc. IEEE/CVF Conf. Comput. Vis. Pattern Recognit. (CVPR)},
  pages = {29707--29716}
}

@article{singh2026pixelup,
  author = {Singh, Deepank and Nihal, Anurag and Hoskere, Vedhus},
  title = {{PixelUp: Zero-shot semantic feature upsampling for fine-grained vision tasks}},
  year = {2026},
  journal = {arXiv preprint arXiv:2608.02792},
  doi = {10.48550/arxiv.2608.02792},
  note = {doi: \href{https://doi.org/10.48550/arxiv.2608.02792}{\nolinkurl{10.48550/arxiv.2608.02792}}}
}

@inproceedings{strudel2021segmenter,
  author = {Strudel, Robin and Garcia, Ricardo and Laptev, Ivan and Schmid, Cordelia},
  title = {{Segmenter: Transformer for semantic segmentation}},
  year = {2021},
  booktitle = {Proc. IEEE/CVF Int. Conf. Comput. Vis. (ICCV)},
  pages = {7242--7252},
  doi = {10.1109/iccv48922.2021.00717},
  note = {doi: \href{https://doi.org/10.1109/iccv48922.2021.00717}{\nolinkurl{10.1109/iccv48922.2021.00717}}}
}

@inproceedings{tomasi1998bilateral,
  author = {Tomasi, C. and Manduchi, R.},
  title = {{Bilateral filtering for gray and color images}},
  year = {1998},
  doi = {10.1109/iccv.1998.710815},
  booktitle = {Proc. 6th Int. Conf. Comput. Vis. (ICCV)},
  pages = {839--846},
  note = {doi: \href{https://doi.org/10.1109/iccv.1998.710815}{\nolinkurl{10.1109/iccv.1998.710815}}}
}

@article{wang2026sav,
  author = {Wang, Xiao and Wang, Ziwen and Wu, Wentao and Wang, Anjie and Wu, Jiashu and Pan, Yantao and Li, Chenglong},
  title = {{Segment Any Vehicle: Semantic and visual context driven SAM and a benchmark}},
  journal = {IEEE Trans. Circuits Syst. Video Technol.},
  year = {2026},
  volume = {36},
  number = {10},
  pages = {15230--15243},
  doi = {10.1109/tcsvt.2026.3708475},
  note = {doi: \href{https://doi.org/10.1109/tcsvt.2026.3708475}{\nolinkurl{10.1109/tcsvt.2026.3708475}}}
}

@inproceedings{wimmer2026anyup,
  author = {Wimmer, Thomas and Truong, Prune and Rakotosaona, Marie-Julie and Oechsle, Michael and Tombari, Federico and Schiele, Bernt and Lenssen, Jan Eric},
  title = {{AnyUp: Universal feature upsampling}},
  year = {2026},
  booktitle = {Proc. Int. Conf. Learn. Represent. (ICLR)},
  url = {https://proceedings.iclr.cc/paper_files/paper/2026/hash/e440aff7f4879b6713867d22c5ec3958-Abstract-Conference.html},
  pages = {140700--140720}
}

@inproceedings{xie2021segformer,
  author = {Xie, Enze and Wang, Wenhai and Yu, Zhiding and Anandkumar, Anima and Alvarez, Jose M. and Luo, Ping},
  title = {{SegFormer: Simple and efficient design for semantic segmentation with transformers}},
  year = {2021},
  booktitle = {Adv. Neural Inf. Process. Syst. (NeurIPS)},
  pages = {12077--12090},
  volume = {34},
  url = {https://proceedings.neurips.cc/paper_files/paper/2021/hash/64f1f27bf1b4ec22924fd0acb550c235-Abstract.html}
}

@inproceedings{yin2025mona,
  author = {Yin, Dongshuo and Hu, Leiyi and Li, Bin and Zhang, Youqun and Yang, Xue},
  title = {{5\%{$>$}100\%: Breaking performance shackles of full fine-tuning on visual recognition tasks}},
  year = {2025},
  booktitle = {Proc. IEEE/CVF Conf. Comput. Vis. Pattern Recognit. (CVPR)},
  pages = {20071--20081},
  doi = {10.1109/cvpr52734.2025.01869},
  note = {doi: \href{https://doi.org/10.1109/cvpr52734.2025.01869}{\nolinkurl{10.1109/cvpr52734.2025.01869}}}
}

@inproceedings{zhou2022ibot,
  author = {Zhou, Jinghao and Wei, Chen and Wang, Huiyu and Shen, Wei and Xie, Cihang and Yuille, Alan and Kong, Tao},
  title = {{iBOT: Image BERT pre-training with online tokenizer}},
  year = {2022},
  booktitle = {Proc. Int. Conf. Learn. Represent. (ICLR)},
  url = {https://openreview.net/pdf/12096c8a70fdac02ab174d59593366e17a8f8e3b.pdf}
}

@inproceedings{ravfogel2020inlp,
  author = {Ravfogel, Shauli and Elazar, Yanai and Gonen, Hila and Twiton, Michael and Goldberg, Yoav},
  title = {{Null it out: Guarding protected attributes by iterative nullspace projection}},
  year = {2020},
  booktitle = {Proc. 58th Annu. Meet. Assoc. Comput. Linguist. (ACL)},
  doi = {10.18653/v1/2020.acl-main.647},
  pages = {7237--7256},
  note = {doi: \href{https://doi.org/10.18653/v1/2020.acl-main.647}{\nolinkurl{10.18653/v1/2020.acl-main.647}}}
}

@inproceedings{belrose2023leace,
  author = {Belrose, Nora and Schneider-Joseph, David and Ravfogel, Shauli and Cotterell, Ryan and Raff, Edward and Biderman, Stella},
  title = {{LEACE: Perfect linear concept erasure in closed form}},
  year = {2023},
  booktitle = {Adv. Neural Inf. Process. Syst. (NeurIPS)},
  doi = {10.52202/075280-2884},
  pages = {66044--66063},
  volume = {36},
  note = {doi: \href{https://doi.org/10.52202/075280-2884}{\nolinkurl{10.52202/075280-2884}}}
}

@article{singh2026ease,
  author = {Singh, Deepank and Nihal, Anurag and Hoskere, Vedhus},
  title = {{Excite, Attend and Segment (EASe): Domain-agnostic fine-grained mask discovery with feature calibration and self-supervised upsampling}},
  year = {2026},
  journal = {arXiv preprint arXiv:2604.00276},
  doi = {10.48550/arxiv.2604.00276},
  note = {doi: \href{https://doi.org/10.48550/arxiv.2604.00276}{\nolinkurl{10.48550/arxiv.2604.00276}}}
}

@inproceedings{ding2026raysup,
  author = {Ding, Yuchuan and Li, Linfei and Zhang, Lin and Shen, Ying},
  title = {{RaysUp: Ultra-light universal feature upsampling via geometry-aware ray representation}},
  year = {2026},
  doi = {10.1007/978-3-032-37132-4_8},
  pages = {130--147},
  booktitle = {Proc. Eur. Conf. Comput. Vis. (ECCV)},
  note = {doi: \href{https://doi.org/10.1007/978-3-032-37132-4_8}{\nolinkurl{10.1007/978-3-032-37132-4_8}}}
}
\end{document}